\documentclass[letterpaper, 10 pt, conference]{ieeeconf}  

\IEEEoverridecommandlockouts                              

\usepackage{mathtools}
\usepackage{amsfonts}  
\usepackage{graphics}
\usepackage[pdftex]{graphicx}
\usepackage{wrapfig}
\usepackage{color}
\usepackage{dsfont}
\usepackage{xparse}

\usepackage[bookmarks=true]{hyperref}
\hypersetup{
colorlinks=true,
urlcolor=blue,
}

\usepackage{algorithmicx}
\usepackage[ruled]{algorithm}
\usepackage{algpseudocode}
\usepackage{amssymb}

\usepackage{hyperref}
\usepackage{amsmath,amsfonts,bm}
\usepackage{bbm}

\algnewcommand\algorithmicforeach{\textbf{for each}}
\algdef{S}[FOR]{ForEach}[1]{\algorithmicforeach\ #1\ \algorithmicdo}

\DeclarePairedDelimiterX{\infdivx}[2]{(}{)}{%
  #1\;\delimsize\|\;#2%
}

\usepackage{expl3}
\ExplSyntaxOn
\newcommand\latinabbrev[1]{
  \peek_meaning:NTF . {
    #1\@}%
  { \peek_catcode:NTF a {
      #1.\@ }%
    {#1.\@}}}
\ExplSyntaxOff
\def\eg{\latinabbrev{e.g}}

\newcommand{\obj}[1]{#1}
\newcommand{\pred}[1]{\texttt{#1}}
\NewDocumentCommand \prop {>{\SplitList{ }} m} {\proposition #1}
\NewDocumentCommand \proposition {g g g g} {(\texttt{#1}~#2
  \IfValueTF{#3}{~#3}{}
  \IfValueTF{#4}{~#4}{}
  )
}
\NewDocumentCommand \action {>{\SplitList{ }} m} {\actioncall #1}
\NewDocumentCommand \actioncall {g g g g} {\text{#1}(#2
  \IfValueTF{#3}{,#3}{}
  \IfValueTF{#4}{,#4}{}
  \texttt{)}
}

\title{Active Task Randomization: Learning Robust Skills via Unsupervised Generation of Diverse and Feasible Tasks}

\author{
    Kuan Fang$^{*1}$,
    Toki Migimatsu$^{*2}$,
    Ajay Mandlekar$^3$,
    Li Fei-Fei$^2$,
    Jeannette Bohg$^2$
    \thanks{$^{*}$Authors have contributed equally.}
    \thanks{$^{1}$University of California, Berkeley, $^{2}$Stanford University, $^{3}$NVIDIA.}
    \thanks{Toyota Research Institute provided funds to support this work.}
}

\def \MethodName {Active Task Randomization}
\def \MethodAcronym {ATR}

\usepackage[noadjust]{cite}

\begin{document}
\maketitle


\begin{abstract}



Solving real-world manipulation tasks requires robots to have a repertoire of skills applicable to a wide range of circumstances.
When using learning-based methods to acquire such skills, the key challenge is to obtain training data that covers diverse and feasible variations of the task, which often requires non-trivial manual labor and domain knowledge.
In this work, we introduce Active Task Randomization (ATR), an approach that learns robust skills through the unsupervised generation of training tasks. ATR selects suitable tasks, which consist of an initial environment state and manipulation goal, for learning robust skills by balancing the diversity and feasibility of the tasks. We propose to predict task diversity and feasibility by jointly learning a compact task representation. The selected tasks are then procedurally generated in simulation using graph-based parameterization. The active selection of these training tasks enables skill policies trained with our framework to robustly handle a diverse range of objects and arrangements at test time.
We demonstrate that the learned skills can be composed by a task planner to solve unseen sequential manipulation problems based on visual inputs.
Compared to baseline methods, ATR can achieve superior success rates in single-step and sequential manipulation tasks. 
Videos are available at \href{https://sites.google.com/view/active-task-randomization/}{sites.google.com/view/active-task-randomization/}

\end{abstract}


\section{Introduction}

\begin{figure}
    \vspace{-3mm}
    \centering
    \includegraphics[width=\linewidth]{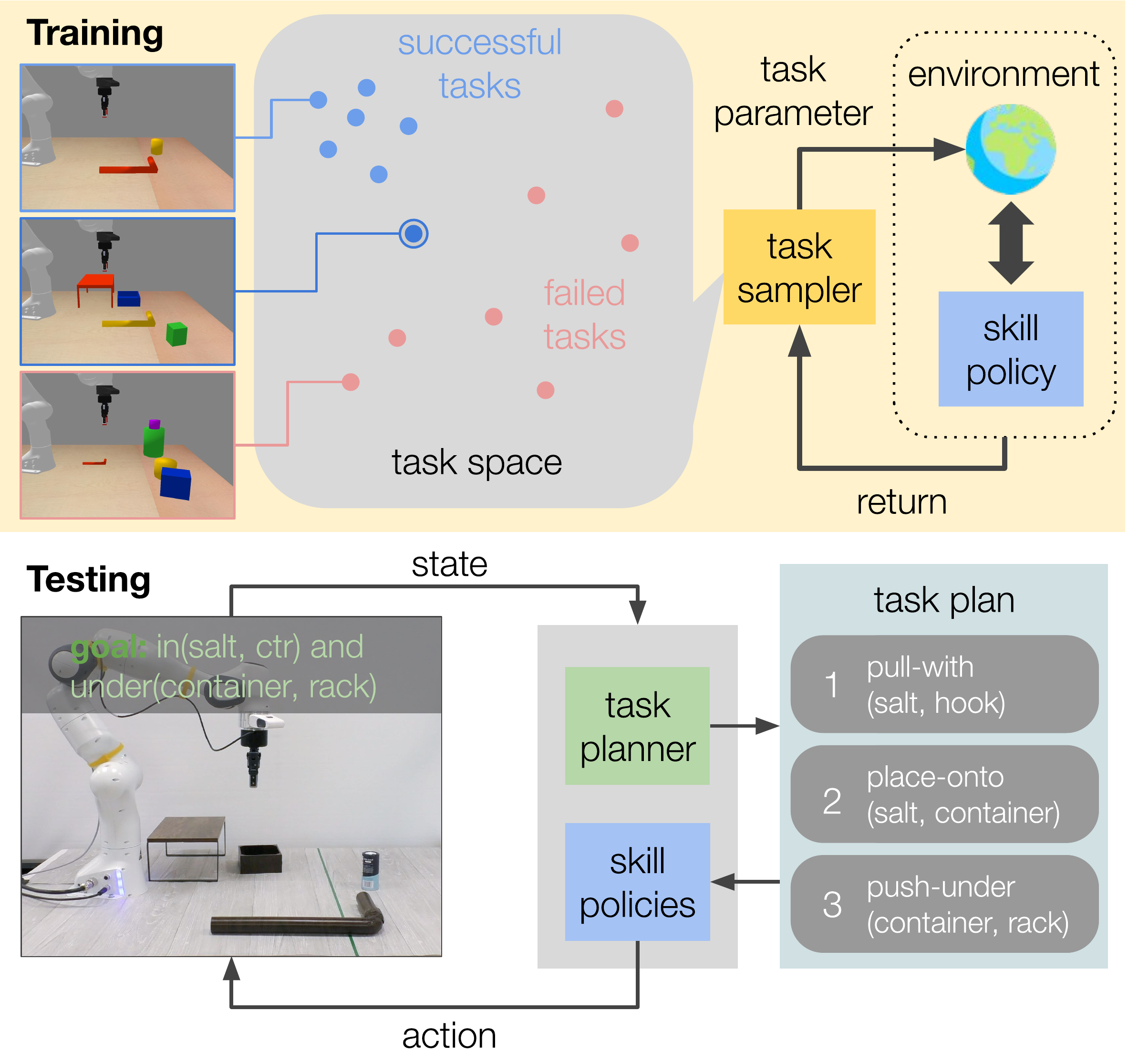}
    \caption{\textbf{Overview of \MethodAcronym.} We use diverse procedurally generated tasks in simulation to train generalizable skills to solve sequential manipulation tasks in the real world. During training, a task sampler proposes suitable tasks in simulation to efficiently train skill policies, by estimating the feasibility and diversity of sampled tasks. During testing, the trained skill policies are composed by a task planner to solve unseen sequential manipulation tasks.
    }
    \label{fig:intro}
    \vspace{-6mm}
\end{figure}

Enabling robots to solve a wide range of manipulation tasks has been a long-standing goal for robotics.
Consider the example in Fig.~\ref{fig:intro}, where the robot is asked to store the salt in a container under the rack. 
Initially, the container is in front of the rack and the salt is outside the robot's workspace.
To accomplish the task, the robot needs to pick up a hook and use it to pull the salt closer, pick up the salt and put it into the container, and finally push the container under the rack.
Each of these skills requires the robot to understand different aspects of the environment based on its raw sensory inputs and manipulate the objects in a way that makes progress toward the desired goal.
Data-driven methods have allowed robots to learn skills such as grasping and pushing from interactions with the physical world~\cite{pinto2015supersizing,agrawal2016poking,levine2017grasping, mahler2017dex}.
However, learning skills that can generalize to unseen environments and goals at test time requires training on interaction data that can cover such variations. 
This training data is usually collected with manually specified environments and goals, which can require non-trivial labor and domain knowledge.
To acquire a repertoire of skills that can tackle the vast complexity and variety of tasks in the real world, we need methods to scale up data collection and learning with minimal human effort. 


One such method to enable policies to handle unseen manipulation tasks in the real world is to randomly generate environments and goals for training in simulation~\cite{Tobin2017DomainRF, OpenAI2019SolvingRC}.
Many of these works have only considered randomizing a limited set of environment properties, such as lighting and textures, for specific domains. 
More challenging domains, such as those involving sequential manipulation, require considering properties such as object types (e.g. salt and rack), configurations (e.g. whether the salt is on or under the rack), and manipulation objectives (e.g. whether the salt should be placed under the rack or the table should be placed on the salt).
The choice of these properties can have a significant impact on \textit{whether} and \textit{how} the task can be solved.
Naive randomization will often result in infeasible tasks or fail to cover the full range of environment and goal configurations that the robot might encounter during test time.
Recently, several works have proposed to adaptively select diverse tasks through curriculum learning and active learning~\cite{Wang2019POETOC, portelas2020teacher, Racanire2020AutomatedCG, fang2020aptgen, Dennis2020EmergentCA, Jiang2021PrioritizedLR}. 
While these works have shown promise in games and grid-world domains, how to effectively leverage adaptive task generation to solve complex real-world robotics problems remains an open challenge.

We present \MethodName~(\MethodAcronym), an approach that learns robust skills through unsupervised generation of training tasks in simulation.
At the heart of our approach is a task sampler that samples candidate tasks for training the skill policies, as shown in Fig.~\ref{fig:intro}.
The sampling is performed in a task space that uses scene graphs to parameterize a wide variety of environment configurations and goals.
To select suitable training tasks, we introduce a ranking score that considers both the feasibility and diversity of the sampled tasks.
Task feasibility is related to the expected return of the current policy. Task diversity is related to the entropy of the task parameters already seen during training. 
To adaptively estimate these two metrics, \MethodAcronym~computes a compact embedding of each task based on the scene graphs using a relational neural network~\cite{santoro2017simple}, which is jointly trained with the skill policies.
Using \MethodAcronym, we can learn various skills (\eg~pushing objects under the rack, pulling objects with tools, and picking and placing objects) without meticulously designing environment initial states and goals for each skill. 
The learned skill policies, which take point clouds as input, are able to generalize to unseen object shapes and arrangements in the real world.
The learned skills can also be composed by a task planner to solve various sequential manipulation tasks.


\textbf{Contributions.}
1) We propose a system that generates training tasks in simulation using a graph-based task parameterization to produce a diverse range of environment configurations and task goals. 
2) We introduce a method that adaptively estimates the feasibility and diversity of generated tasks through a compact task embedding learned jointly with the skills.
3) We demonstrate that the skills learned using our method can solve unseen single-step and sequential manipulation tasks based on visual inputs.

\begin{figure*}[t!]
    \centering
    \includegraphics[width=\linewidth]{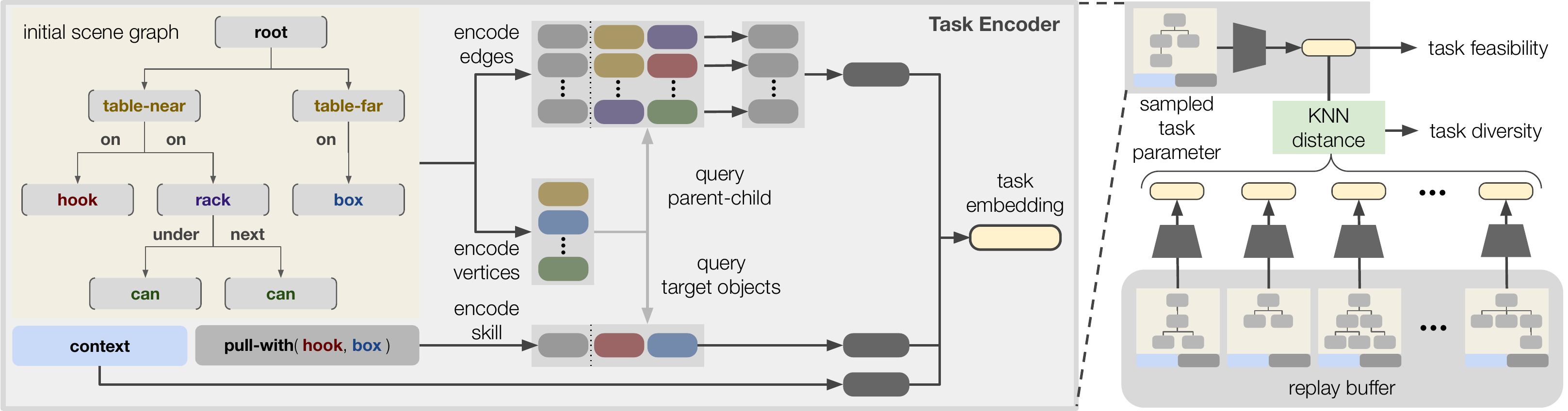}
    \caption{
    \textbf{\MethodAcronym~Architecture.} We use a task encoder (left) to project each task parameter $w$ into a compact embedding for estimating the feasibility and diversity of sampled tasks (right). The task encoder is implemented with a relational neural network that fuses the information of the initial scene graph, the planned action, and the context variables. Given the computed task embeddings, the task's feasibility is the expected return estimated by a learned value function, and its diversity is the negative log density of the sampled task parameter estimated through non-parametric density estimation.
    }
    \vspace{-3mm}
    \label{fig:model}
\end{figure*}

\section{Related Work}

\subsection{Skill Learning}

Skill learning has been widely studied in robotics. 
%
Given large-scale datasets, a variety of skills such as grasping, pushing, rotating, and inserting can be learned~\cite{
mahler2017dex, Dasari2019RoboNetLM, 
mandlekar2018roboturk, mandlekar2019scaling, Ebert2022BridgeDB}.
In well-designed simulated and real-world environments~\cite{James2020RLBenchTR, yu2019meta, fang2019cavin, corl2018surreal, levine2017grasping, pinto2015supersizing, kalashnikov2021mtopt}, the skills can also be learned in a self-supervised manner, in which the trained policy is used to continuously collect data to bootstrap its performance.
%
%
Although these works have shown promising results towards learning skills, manually scaling up the datasets and environments to cover the vast diversity and complexity of real-world scenarios is impractical. 
Instead of training skills on hand-designed environments, we propose training them on a diverse set of environments that are adaptively generated based on the current proficiency of the skills, which facilitates generalization to new scenarios.

\subsection{Procedural Generation for Robotics}
There have been an increasing number of works that use procedural content generation for robotics~\cite{summerville2018procedural, risi2019procedural, Cobbe2019LeveragingPG}. 
Several works use procedurally generated environments in simulation with randomized object properties and dynamics~\cite{Tobin2017DomainRF, Peng2018SimtoRealTO}. 
More complex environments can also be generated using pre-defined probabilistic programming languages~\cite{Innes2021ProbRobSceneAP, Innes2021AutomaticSO, Chamzas2022MotionBenchMakerAT}.
Most of these works use hand-designed procedural generation processes, which requires non-trivial expertise and labor.

Recent works aim to learn generative models to capture the distribution of realistic environments, goals, and tasks given previously collected datasets~\cite{kim2020learning, Kim2021DriveGANTA, Paschalidou2021ATISSAT}.
Given a target task, \cite{OpenAI2019SolvingRC} and \cite{Mehta2019ActiveDR} actively randomize the hyperparameters of the simulated environments to facilitate domain adaptation. 
However, large-scale datasets or target environments are often unavailable beforehand and data distributions chosen by human experts might not always reflect the suitable difficulty and diversity required by the learning algorithm.

Other works adaptively generate environments and tasks as curricula through the interplay between the trained policy and an adversarial policy~\cite{Sukhbaatar2017IntrinsicMA, Dennis2020EmergentCA, Sukhbaatar2018LearningGE} or a generator~\cite{Forestier2017IntrinsicallyMG, Held2018AutomaticGG, Racanire2020AutomatedCG, Lee2020LearningQL, Zhang2020AutomaticCL, Khalifa2020PCGRLPC, Bontrager2021LearningTG, fang2020aptgen, fang2021slide}.
\cite{Wang2019POETOC, Wang2020EnhancedPO, Jiang2021PrioritizedLR, ParkerHolder2022EvolvingCW, Fontaine2020AQD, Fontaine2021OnTI} propose to reuse and evolve previously learned environments to learn increasingly difficult and diverse skills.
While these methods have shown success in grid-world or purely simulated domains, real-world robotics domains still pose an unsolved challenge because of the large space of possible environments and tasks. Naively generating random environments for robotics domains is unlikely to result in feasible tasks or cover the vast diversity of environment configurations that a robot may experience at runtime. We propose a method to generate tasks using a metric that considers both task feasibility and diversity with a graph-based representation that is learned jointly with the policies. This metric allows our procedural generation method to generate open-ended robotics tasks in environments involving multiple objects and skills.

\subsection{Sequential Manipulation}

Sequential manipulation tasks present a challenging problem for generalizable skill learning due to their long horizons, large exploration space, and infinite combinations of initial states and goals. Traditional Task and Motion Planning (TAMP) algorithms~\cite{garrett2021integrated,kaelbling2010hierarchical,toussaint2018differentiable,garrett2020pddlstream,migimatsu2020objectcentric} can generalize to any sequential manipulation task through the use of a task planner, but they require hand-crafted policies with full knowledge of the environment state and dynamics.
While these limitations can be addressed by learning, works that seek to replace hand-crafted policies with learned policies for sequential manipulation offer limited generalization to unseen tasks~\cite{da2012learning,masson2016reinforcement,chitnis2020efficient,lee2021adversarial,dalal2021accelerating,xu2021deep,nasiriany2022maple}. Generalizing to a wide variety of sequential manipulation tasks requires training policies on diverse environments and tasks during training \cite{oh2017zero,xu2019regression,driess2021learning}. Our framework can be applied to sequential manipulation to provide the diversity of feasible tasks required for training generalizable skills.

\section{Problem Formulation}
\label{sec:background}

\MethodAcronym~seeks to enable robots to solve a wide variety of real-world manipulation tasks, including tasks that may consist of several stages of manipulation. To enable generalization across a broad set of task initializations and goals, we aim to learn a set of skills and then compose them to solve tasks. In this work, we focus on learning these skills and assume access to a high-level task planner that can sequence the skills. This section provides further details on our problem formulation. First, we describe the structure of the set of skills that \MethodAcronym~uses and how it interfaces with the task planner to solve tasks (Sec.~\ref{sec:skills}). Next, we describe the structure of policies that are trained for each skill (Sec.~\ref{sec:policies}). Finally, we describe how tasks are parameterized so that a learned module can generate tasks for policies to solve during training (Sec.~\ref{sec:task_generation}). 




\subsection{Skills}
\label{sec:skills}

A manipulation task consists of an initial state and a task goal. For example, the bottom of Fig.~\ref{fig:intro} shows a task where several items begin on a table surface. The task goal is for the robot to store the salt can in a container under the rack. We assume that such tasks can be decomposed into a sequence of skills by using a task planner. Each skill is associated with a symbolic action defined by its symbolic pre- and post-conditions, specified using PDDL~\cite{mcdermott1998pddl}. This allows the task planner to take in the current symbolic state $G$ and propose a sequence of skills that can achieve the given task goal. In the example in Fig.~\ref{fig:intro}, the task planner produces a plan with three skills: $\action{pull-with i j}$, $\action{place-onto i j}$, and $\action{push-under i j}$. 

The symbolic state $G$ is a scene graph $(O, E)$, where $O$ is the set of objects in the scene and $E$ is the set of spatial relationships between them; each object $o \in O$ represents one node and each spatial relationship $e \in E$ represents a directed edge between two nodes. For example, at the start of the task in Fig.~\ref{fig:intro}, the hook is on the table, so $E$ would contain a directed $\pred{on}$ edge from $\obj{table}$ to $\obj{hook}$. In PDDL language, this represents the proposition $\prop{on hook table}$. We assume that a method for perceiving the current scene graph of the environment is given.
We represent the skills output by the task planner as tuples $c = (k, i, j)$, where $k$ is the skill, and $i$ and $j$ are the target objects that the robot should manipulate. For example, the skill $\action{pull-with salt hook}$ would have $k = \text{pull-with}$, $i = salt$, and $j = hook$.

While the task planner produces the sequence of skills and the objects involved in each skill, it does not specify the precise robot controls needed to carry out each skill successfully. Consequently, each skill additionally consists of a learned policy. The policy is meant to output parameters for the manipulation primitive in each skill. For example, the $\action{place-onto i j}$ primitive is parameterized by parameters $a$ consisting of a pick pose to pick object $i$ and placement pose of $i$ on $j$. The primitive outputs the control commands to execute a trajectory for picking and placing $i$ on $j$ in a single time step. The corresponding policy will take in the current environment state and output primitive parameters (i.e. action) $a$.
Note that skill parameters are often non-trivial to specify without the use of learning-based methods, especially to ensure that skills are functional for a wide variety of objects and scenes. For example, picking up the $\obj{hook}$ requires understanding the geometry of its handle and head; simply picking it up at the center of its bounding box results in grasping empty space.

\subsection{Policies}
\label{sec:policies}
Given a library of manipulation primitives, our objective is to learn policies for each skill that can be sequenced together at test time to solve real-world manipulation tasks. A policy $\pi_k(a | s, c)$ for skill $k$ takes as input the current environment state $s$ along with skill context $c = (k, i, j)$ and outputs parameters $a$ for the corresponding primitive. The state $s = (x, O, E)$ consists of a point cloud $x$, list of objects $O$, and spatial relationships $E$ between the objects. Each object $o \in O$ is accompanied by a segmentation mask of the object in the point cloud. We assume that a method for segmenting point clouds is given.


We train the policies in a self-supervised manner. During training, a policy $\pi_k$ receives the current state $s$ and context $c$ and outputs primitive parameters $a$. The policy receives a binary reward $r = R_k(s', c)$, which indicates when the success condition defined for skill $k$ is satisfied at the resulting state $s'$. We store the collected experience $(a, s, c, r)$ in a replay buffer $\mathcal{D}$.
The policies are trained to imitate the successful trajectories by maximizing the mean log likelihood $\log \pi_k(a | s, c)$ for $(a, s, c, r) \in \mathcal{D}$ with $r = 1$.

\subsection{Task Generation}
\label{sec:task_generation}

We are interested in training policies that can be used to solve a diverse range of tasks; thus, the policies need to be exposed to a diverse range of tasks during training. To do so, we leverage a black-box procedural task generation module $M$ that instantiates tasks from task parameters $w$ as in \cite{portelas2020teacher, fang2019cavin}. We propose a graph-based task parameterization for manipulation tasks in Sec.~\ref{sec:task-parameterization}.

To generate tasks for training the policies, we sample $w$ from a task parameter distribution $p(w)$ and feed it to the task generation module to get a task $M(w)$. In the simplest case, $p(w)$ can be a uniform distribution over the task parameter space. In Sec.~\ref{sec:task-diversity-feasibility}, we propose learning a distribution that balances task feasibility and diversity to train policies that can generalize better to unseen tasks.




\section{\MethodName}
\label{sec:method}

Our approach, \MethodName~(\MethodAcronym), learns skill policies for solving real-world manipulation tasks through unsupervised generation of tasks in simulation.
Real-world manipulation tasks can encompass a broad range of scenarios. To cover this space, a task selection mechanism would require selecting from a large set of object properties, arrangements, and goals. Furthermore, naive task selection strategies can result in task parameters that violate physical constraints or goals that are infeasible.

At the heart of our method is a task sampler that generates tasks for training the skill policies (Fig.~\ref{fig:intro}).
In each episode, the task sampler samples a task parameter $w$ from a distribution $p(w)$ and then selects the optimal task parameter based on the prior experience of the robot. 
The selected task parameter is then used to instantiate a training task in simulation through the procedural generation module $M$ described in Sec.~\ref{sec:task_generation}.
The created task is then used for data collection and training the skill policy.


In this section, we first introduce a graph-based parameterization of tasks that enables procedural generation of a wide range of possible environments and goals. Then we describe a procedure that adaptively proposes diverse and feasible tasks. Finally, we explain how to learn compact task embeddings to facilitate the estimation of feasibility and diversity of sampled tasks.

\subsection{Graph-Based Parameterization of Manipulation Tasks}
\label{sec:task-parameterization}

We aim to develop a task parameterization that is expressive enough to represent a wide range of environments and goals.
Inspired by \cite{portelas2020teacher, fang2020aptgen}, we define a parameter space $\mathcal{W}$ that represents the inter-task variation of the task space $\mathcal{T}$ (\eg~the symbolic relationships of the objects, the object types, and the object sizes) and use a predefined black-box procedural generation module $M$ to instantiate each task $M(w)$.

To represent the object properties and symbolic relationships between objects, we devise a graph-based parameterization that can be converted to PDDL~\cite{mcdermott1998pddl}. 
Each manipulation task is defined as a tuple $w = (O, E, C, u)$, where $O$ is a list of objects (i.e. scene graph nodes) with each object defined by its intrinsic properties like type and size that do not change across time, $E$ is a list of spatial object relationships (i.e. scene graph edges) in the initial state, $C$ is a list of skill contexts $c$ that specify the skills to be executed, and $u$ contains environment context such as the camera pose. 
These task parameters $w$ are sampled from a domain-specific prior probability distribution $p(w)$.

The most naive task generation strategy would be to uniformly sample $w$. However, sampling $w$ randomly would be unlikely to result in a feasible task. For example, a randomly sampled infeasible task could be placing the table on the rack. Another infeasible example could be pushing a can under the rack, where the can is too large to fit. The need to sample feasible tasks while providing a diverse set of training tasks for policies motivates our method, which selects tasks based on estimates of their feasibility and diversity.

\subsection{Estimation of Task Feasibility and Diversity}
\label{sec:task-diversity-feasibility}


Task feasibility gauges if a task specified by $w$ can be solved by the current policy $\pi$. Feasibility evolves during training, as a task which is infeasible now may become feasible as the policy learns.
Following \cite{fang2020aptgen}, we define task feasibility as the expected return $\mathbb{E}_{\pi}[ r | w ]$ of unrolling the policy $\pi$ in the task specified by $w$. 
We train a value function $V(w)$ to estimate the expected return by minimizing the mean squared error
\begin{eqnarray}
    \mathcal{L} = \mathbb{E}_{r, w \sim \mathcal{D}} \left[ || V(w) - r ||^2 \right].
    \label{eqn:value_objective}
\end{eqnarray}

Inspired by recent work on unsupervised exploration~\cite{liu2021behavior}, we measure the task diversity of a sampled $w$ through entropy estimation. 
Specifically, let $p(\cdot)$ be the density of task parameters that the policy has been trained on. Then, task diversity is defined as the entropy $\mathbb{E}[-\log p(w)]$. 
While directly computing the density $p(w)$ is intractable, we estimate $\hat{p}(w)$ through particle-based estimation~\cite{beirlant1997nonparametric, jiao2018nearest}.
For each sampled $w$, we denote $w_K$ to be the $K$-th nearest neighbor (KNN) of $w$ in the set of all task parameters stored in the replay buffer $\mathcal{D}$. The distance between two task parameter vectors is computed with a distance metric $d(\cdot, \cdot)$, an example of which is described in Sec.~\ref{sec:task_embedding}. The particle-based approximation of the task parameter density can then be computed as
\begin{eqnarray}
     \hat{p}(w) 
    = - \frac{K}{m V_K},
\end{eqnarray}
where $V_K$ is the volume of the hyper-sphere of radius $d(w, w_K)$. 
To encourage the training tasks to be diverse, we would like each training task $w$ to have a high distance $d(w, w_K)$. Intuitively, this can be considered to be searching for the $w$ that is most distinguishable from previous task parameters stored in $\mathcal{D}$. Following the practice of prior work~\cite{liu2021behavior}, we compute the nearest neighbors $w_K$ from a sampled subset from $\mathcal{D}$ rather than the entire dataset for computational efficiency.

At the start of each episode, we sample a batch of task parameters and compute the score for each $w$ as
\begin{eqnarray}
    f(w) = V(w) + \beta d(w, w_K),
    \label{eqn:ranking_score}
\end{eqnarray}

where $\beta$ is a weight that balances the two terms and is chosen to be 0.1 based on grid search. This score is used to construct a categorical distribution with logits $f(w)$, which can be used to sample $w$. To encourage exploration, we use an epsilon-greedy strategy that directly samples from the prior distribution $p(w)$ with 10\% probability and from the categorical distribution otherwise.

\subsection{Learning Compact Task Embeddings}
\label{sec:task_embedding}

We learn a compact representation of the task parameter $w$ to effectively estimate task feasibility and task diversity.
To extract information from the heterogeneous task parameter $w$, we design a task encoder $\phi(w)$ (as shown in Fig.~\ref{fig:model}) with a relational network~\cite{santoro2017simple} that is jointly learned with the skill policies.
The task encoder first separately encodes the scene graph $(O, E_{\text{init}})$, the environment context $u$, and the skill contexts $C$ into three vectors using a set of sub-level encoders, and then concatenates these vectors to compute the compact task encoding. 
For the scene graph, each vertex $o \in O$ (object attributes) and each edge $e \in E$ (spatial relationships) are separately encoded as $\phi_o(o)$ and $\phi_e(e)$ where the vertex encoder $\phi_o$ and edge encoder $\phi_e$ are implemented as a single fully-connected (FC) layer.
Then the vertex and edge embeddings are merged by the relational network according to the scene graph.
To represent the skill contexts $C$, we encode each skill index $k$ as $\phi_k(k)$ and encode the target objects $o_i$ and $o_j$ as $\phi_o(o_i)$ and $\phi_o(o_j)$, using the encoders $\phi_k$ and $\phi_o$.
The environment context is concatenated and encoded using a single fully-connected (FC) layer. 
Using the collected trajectories in the replay buffer, $\phi$ is trained through the objective function in Eq.~\ref{eqn:value_objective}.
Given the computed task embedding $\phi(w)$, we can now re-define $V(w)$ and $d(w, w_K)$ using the encoded $\phi(w)$ and $\phi(w_K)$ instead of the original $w$ and $w_K$ as inputs. 
Specifically, we re-write Eq.~\ref{eqn:ranking_score} as
\begin{eqnarray}
    f(w) = V(\phi(w)) + \beta d(\phi(w), \phi(w_K)),
\label{eqn:ranking_score_encoded}
\end{eqnarray}
where $d(\cdot, \cdot)$ is the Euclidean distance and $V(\cdot)$ is implemented as two  fully-connected (FC) layers on top of the computed $\phi(w)$. 
\section{Experiments}
\label{sec:experiments}

\begin{figure*}[t!]
    \centering
    \includegraphics[width=\linewidth]{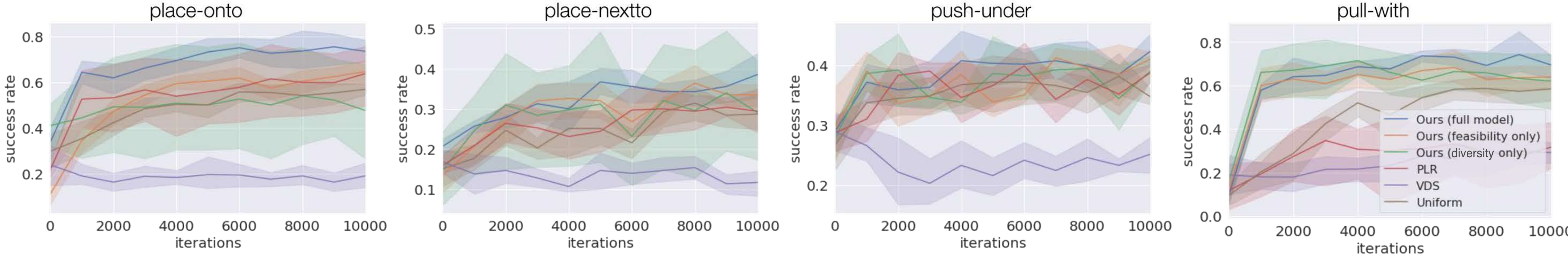}
    \caption{\textbf{Skill Learning Performance.} The average success rates across five training runs is shown with the shaded region indicating the standard deviation. The x-axis indicates the number of collected simulation steps. Our method achieves comparable or better performances for all four skills. }
    \vspace{-3mm}
    \label{fig:skill-learning-performance}
\end{figure*}

\begin{figure}[t!]
    \centering
    \includegraphics[width=\linewidth]{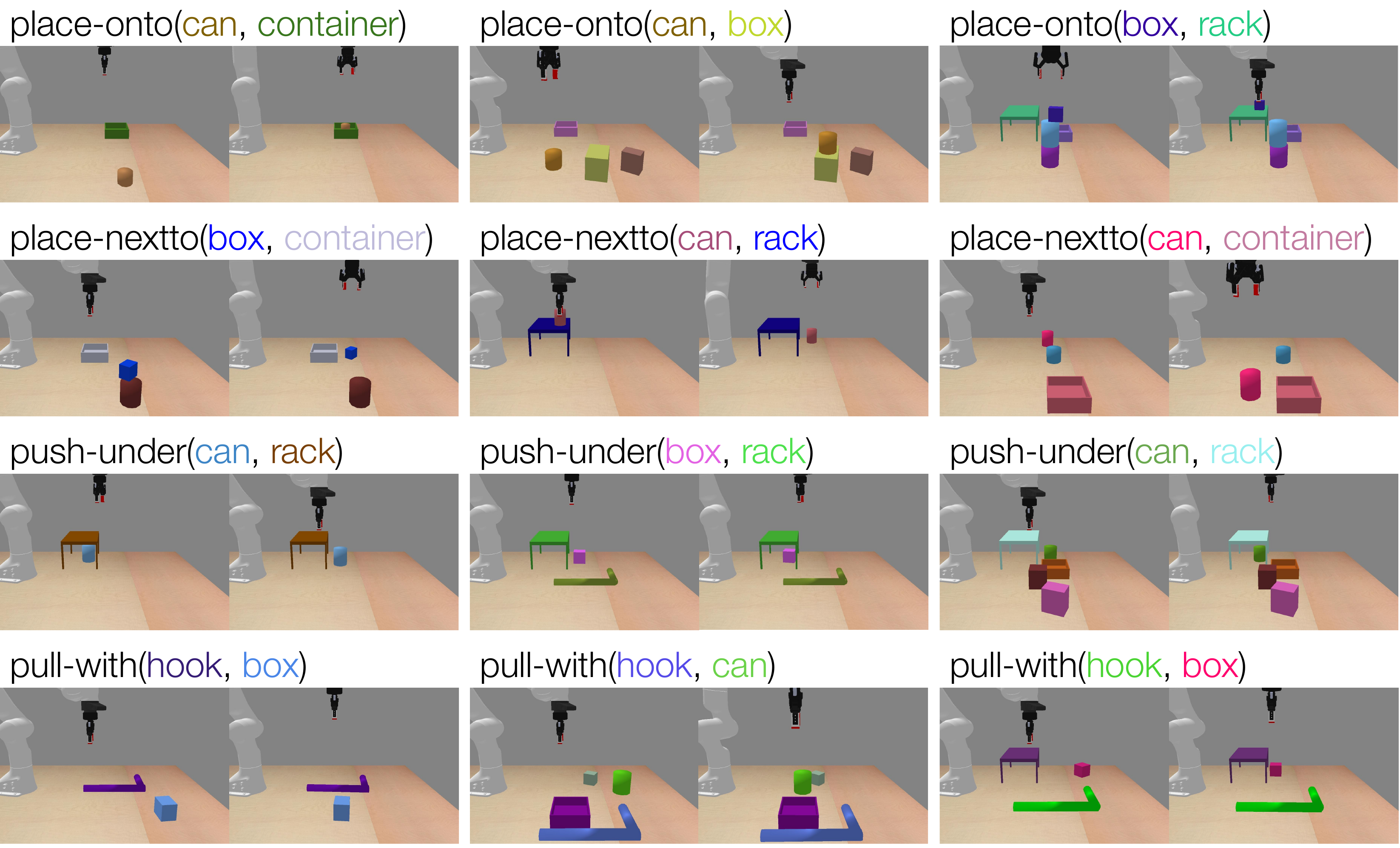}
    \caption{\textbf{Examples of Proposed Tasks.} Tasks with various object types, scales, symbolic relationships, and camera configurations are proposed in simulation. Most of the proposed tasks are feasible and can be solved by the planned actions. The names of the target objects for each skill are color-coded with the same color as they appear in the simulated environment.
    }
    \vspace{-6mm}
    \label{fig:proposed-tasks}
\end{figure}

\begin{figure*}[t!]
    \centering
    \includegraphics[width=\linewidth]{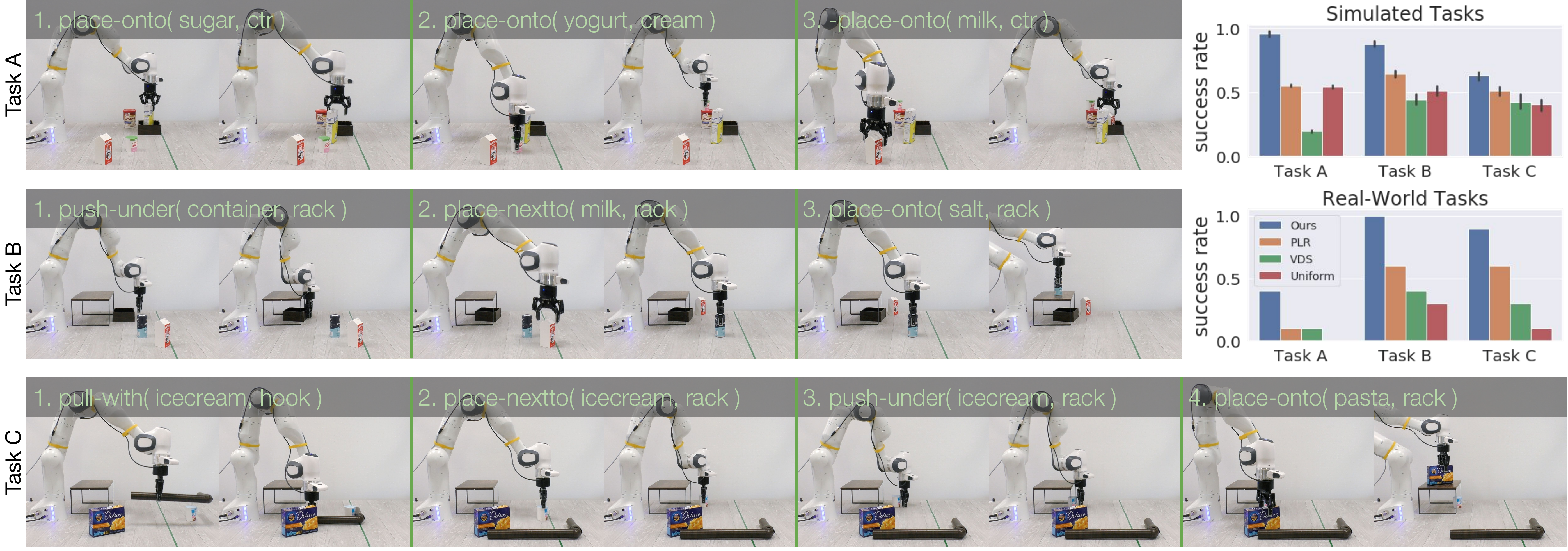}
    \caption{\textbf{Sequential Manipulation Performance.} The video sequences on the left show qualitative results of ATR on the evaluation tasks, and the bar plots on the right show quantitative comparisons with the baselines. The three sequential manipulation tasks require different combinations of the skills and are used for evaluation in both simulation and in the real world. The initial configurations of the objects are randomized for each trial. Using the skills learned using \MethodAcronym, the robot successfully completes the three evaluation tasks with success rates ranging from 88.6--95.8\% in simulation and 40--100\% in the real world. The error bars indicate the standard deviation of the success rates in simulation
    }
    \vspace{-4mm}
    \label{fig:sequential-manipulation-performance}
\end{figure*}

 In our experiments, we seek to answer the following questions: 1) Can \MethodAcronym~procedurally generate feasible and diverse tasks during training? 2) Can the tasks proposed by \MethodAcronym~improve skill learning performance? 3) How well can the learned skills perform in unseen manipulation tasks? 


\subsection{Experimental Setup}

Our experiments are conducted in a real-world table-top manipulation environment, in which a 7-DoF Franka Emika robot arm interacts with various objects on the table. These objects include $\obj{sugar}$, $\obj{yogurt}$, $\obj{cream}$, $\obj{milk}$, $\obj{ice cream}$, $\obj{pasta}$, $\obj{salt}$, $\obj{hook}$, and $\obj{rack}$.

\textbf{States.}
As described in Sec.~\ref{sec:policies}, the environment state that is given to the policies is represented by a tuple $s = (x, O, E)$. An externally mounted Kinect2 camera provides point cloud observations $x$, and segmentations of the objects $O$ in the point cloud are obtained using an external module that segments objects using color thresholds. Spatial relationships $E$ in the scene graph are constructed from these segmented point clouds using hand-crafted heuristics for spatial relationships. For example, the $\prop{on yogurt rack}$ is detected if the bounding box of $\obj{yogurt}$ is above and supported by the bounding box of $\obj{rack}$. We leave the integration of learned segmentation methods and learned scene graph classifiers such as \cite{migimatsu2022grounding} to future work. 

\textbf{Skills.}
This domain contains four skills, $\{\action{place-onto i j}$, $\action{place-nextto i j}$, $\action{push-under i j}$, $\action{pull-with i j}\}$, which each involves two target objects, $i$ and $j$. A policy $\pi_k(a | s, c)$ is trained for each skill $k$ to predict actions $a$ from the environment state $s$ and skill context $c = (k, i, j)$. Each skill comes with a 6-d action space, where actions $a$ are composed of two 3d positions $(p_i, p_j)$ representing end-effector positions relative to the corresponding target object. These actions are provided as input to manipulation primitives which are implemented with closed-loop Operational Space controllers~\cite{khatib1987unified}. For example, the primitive for $\action{place-onto i j}$ will first move the end-effector to position $p_i$ defined relative to object $i$, grasp $i$, move to position $p_j$ defined relative to object $j$, and then release $i$. This action is successful if $i$ is placed stably on top of $j$.

The skills are trained in simulation (using Bullet~\cite{coumans2021}) on tasks generated by our framework. We then evaluate the skills in both simulation and the real world.

\textbf{Task Parameterization.}
As described in Sec.~\ref{sec:task-parameterization}, we generate tasks from sampled task parameters $w = (O, E, C, u)$, where $O$ is the set of objects in the scene, $E$ is the list of spatial relationships between the objects, $C$ is the sequence of skills to execute, and $u$ is the environment context. Each object $o \in O$ is described by its type (rack, container, hook, box, or can), and size. The environment context $u$ contains the camera pose and a parameter which adds noise to the simulated point cloud. This environment context allows the creation of tasks with varying camera poses and noise characteristics, which helps the policies transfer better to the real world.

We represent $w$ as a vector of length 54, which supports up to six objects in the scene (including the table). This vector includes a bit mask of length 6 indicating whether each of the 6 objects appears in the scene.

\textbf{Baselines.}
We compare our method with three domain randomization and task generation baselines and evaluate the success rates. \textbf{Uniform} naively creates random environments by directly sampling from the parameterized task space without considering the feasibility or diversity of the tasks. \textbf{VDS}~\cite{Zhang2020AutomaticCL} uses disagreement among an ensemble of task value networks as a metric of epistemic uncertainty. This is then used to select the optimal task among the sampled candidates. The original VDS is designed for sampling 2D goal coordinates and uses a simple MLP to compute the value, which is unlikely to learn structured representations for high-dimensional heterogeneous task parameters. Instead, we upgrade VDS with task value networks defined in Sec.~\ref{sec:method}. \textbf{PLR}~\cite{schaul2016prioritized} re-samples previous tasks stored in the replay buffer based on a scoring function which measures the learning potential of replaying the task in the future. We also analyze two ablations of our method which consider only feasibility or only diversity when sampling tasks. All methods use the same policy learning algorithm and hyperparameters for a fair comparison.




\textbf{Implementation Details.} 
Adapted from \cite{fang2019cavin}, the policy network takes the segmented point cloud and the target object indices as inputs and produces the mean and standard deviation of the predicted positional offsets.
The task value network first uses 16D fully-connected (FC) layers to embed each vertex and each edge and then merges the information using a 64D FC layer.
Each method is trained for 10,000 training iterations.
In each iteration, one environment step is collected from the simulated environment.
For all experiments, we use a learning rate of $3 \times 10^{-4}$ and a batch size of 128.
We use closed-loop task planning to recover from skill failures. If recovery from a failure is impossible (\eg~an object falls over and rolls away) or the time runs out, we consider the task to be a failure.

\subsection{Proposed Tasks}

Examples of proposed tasks in simulation are shown in Fig.~\ref{fig:proposed-tasks}.
During training, our method successfully proposes tasks with various object types, shapes, and symbolic relationships.
In each task, there exists a feasible planned action to achieve the goal state by interacting with the target objects. 
In contrast, using baseline methods or uniformly sampling from the task space often results in infeasible tasks such as grasping objects that are beyond reach, pulling distant objects with a hook that is too short, pushing objects that are too large to fit beneath the rack, and stacking too many objects on a small surface.

\subsection{Skill Learning}

During training, we evaluate the performance of the skill policies in simulation.
Every 1,000 training iterations, each skill policy is evaluated on five hand-designed single-step tasks that need to be solved by the trained skill for 50 episodes. 
For all four skills, our full \MethodAcronym~method achieves success rates that are comparable or superior to baselines and ablations. 
The average success rates evaluated across five random seeds are presented in Fig.~\ref{fig:skill-learning-performance}, in which the standard deviation is denoted by the shaded region.
The baseline methods often fail to find tasks with valid configurations and their exploration of the task space is less efficient.
The highest performance improvements are achieved in place-onto and pull-with since these two skills have the most stringent requirements for the types and scales of the two target objects. 
The performance of \MethodAcronym~depends on taking both the feasibility and diversity of the task into consideration; using only one of the two terms leads to worse performance.

\subsection{Sequential Manipulation}

For evaluating the trained policies on sequential manipulation tasks, we use a task planner to decide the sequence of skills. The sequential manipulation task contains a goal scene graph, $G_{\text{goal}}$, and the task planner finds a sequence of skills $C$ that reaches the goal scene graph from the current scene graph $G$. Task planning is run in a closed loop every time a skill is executed to recover from potential skill failures. We consider a task successful if it can be completed within 10 steps. 
In simulation, we evaluate each target task with five random seeds for 1,000 episodes per seed. 
In the real world, each task is evaluated for 10 episodes.
The success rates are presented in Fig.~\ref{fig:sequential-manipulation-performance}.
As shown in the bar charts, our \MethodAcronym~method outperforms baselines on all three evaluation tasks in both simulation and in the real world, given the superior performance of individual learned skills.
The baseline methods do not train the skill policies on a diverse enough range of objects and thus often cannot complete the tasks within 10 steps.

\section{Discussion and Future Work}

We presented \MethodAcronym, a method that adaptively proposes feasible and diverse tasks from a parameterized task space to learn robust skills for real-world manipulation. 
Our experiments demonstrated that \MethodAcronym~ can effectively estimate task feasibility and diversity to generate suitable tasks for training a library of skills.
In addition to solving single-step tasks, the learned skills can also be composed by a task planner to solve sequential manipulation tasks.

One limitation of \MethodAcronym~is that the set of skills is predefined. We intend to extend the current method to enable unsupervised skill discovery.
Another limitation is that the task planner is hand-defined and requires an external segmentation pipeline, which could be replaced with a learned high-level policy or a large language model in future work.






\end{document}